\documentclass[lettersize,journal]{IEEEtran}
\usepackage{amsmath,amsfonts}
\usepackage{algorithmic}
\usepackage{algorithm}
\usepackage{array}
\usepackage[caption=false,font=normalsize,labelfont=sf,textfont=sf]{subfig}
\usepackage{textcomp}
\usepackage{stfloats}
\usepackage{url}
\usepackage{verbatim}
\usepackage{graphicx}
\usepackage{cite}
\usepackage{multirow}
\usepackage{makecell}
\usepackage[justification=centering]{caption}

\begin{document}

\title{Pheno-Robot: An Auto-Digital Modelling System for In-Situ Phenotyping in the Field} 

\author{Yaoqiang Pan$^{1,*}$, Kewei Hu$^{1,*}$, Tianhao Liu$^{2}$, Chao Chen$^{2}$, and Hanwen Kang$^{2,\#}$
\thanks{$^{*}$ Equal Contribution}        
\thanks{$^{1} $ K.Hu and Y.Pan are with the College of Engineering, South China Agriculture University, Guangzhou, China}
\thanks{$^{2} $ T.Liu, C.Chen, and H.Kang are with the Department of Mechanical and
Aerospace Engineering, Monash University, Melbourne, Australia}}

\markboth{Journal of \LaTeX\ Class Files,~Vol.~14, No.~8, August~2021}%
{Shell \MakeLowercase{\textit{et al.}}: A Sample Article Using IEEEtran.cls for IEEE Journals}


\maketitle

\begin{abstract}
Accurate reconstruction of plant models for phenotyping analysis is critical for optimising sustainable agricultural practices in precision agriculture. 
Traditional laboratory-based phenotyping, while valuable, falls short of understanding how plants grow under uncontrolled conditions. 
Robotic technologies offer a promising avenue for large-scale, direct phenotyping in real-world environments.
This study explores the deployment of emerging robotics and digital technology in plant phenotyping to improve performance and efficiency. Three critical functional modules: environmental understanding, robotic motion planning, and in-situ phenotyping, are introduced to automate the entire process.
Experimental results demonstrate the effectiveness of the system in agricultural environments. The pheno-robot system autonomously collects high-quality data by navigating around plants. In addition, the in-situ modelling model reconstructs high-quality plant models from the data collected by the robot.
The developed robotic system shows high efficiency and robustness, demonstrating its potential to advance plant science in real-world agricultural environments.
\end{abstract}

\section{Introduction}
\IEEEPARstart{P}{lant} phenotyping plays a fundamental role in precision agriculture. It involves identifying and selecting genetics with advantageous input traits and complementary output traits \cite{fu2020application}. 
While phenotyping in the laboratory plays a key role in identifying promising lines for crossbreeding, they are a surrogate for the primary goal of understanding how a crop will grow in real-world environments \cite{wu2020mvs}.
Uncontrolled 'non-laboratory' conditions present significant challenges, particularly in the analysis of the traits responsible for beneficial responses \cite{deery2021field}. 
Therefore, the mass collection of phenotypic data in the field is essential for precision agriculture.

At the heart of plant phenotyping is the digital modelling of plant growth and traits, including both appearance and geometry \cite{zhang2018imaging}.
This process involves monitoring the growth, development, and changes in plants, taking into account factors such as climate, soil properties, pests, and diseases \cite{xu2022modular}.
Image-based plant modelling is a widely used approach, particularly for analysing observable traits \cite{zhang2018imaging}.
However, analysing plants from a single viewpoint poses challenges, especially when plants overlap \cite{dellen2015growth}. 
To overcome this limitation, RGB images acquired from multiple viewpoints are widely used.
Recently, 3D reconstruction technology has become a prominent tool for plant analysis \cite{feng2021comprehensive}.
This new paradigm provides essential information about the geometric aspects of plant \cite{paulus2019measuring}, including height \cite{virlet2016field}, volume \cite{paulus2014high}, and even mass \cite{xiong2016leaf}. 
These advances in modelling technology, coupled with the increasing availability of information-rich data, contribute significantly to the capture of detailed plant characteristics. 

\begin{figure}[h]
    \centering
    \includegraphics[width=0.95\linewidth]{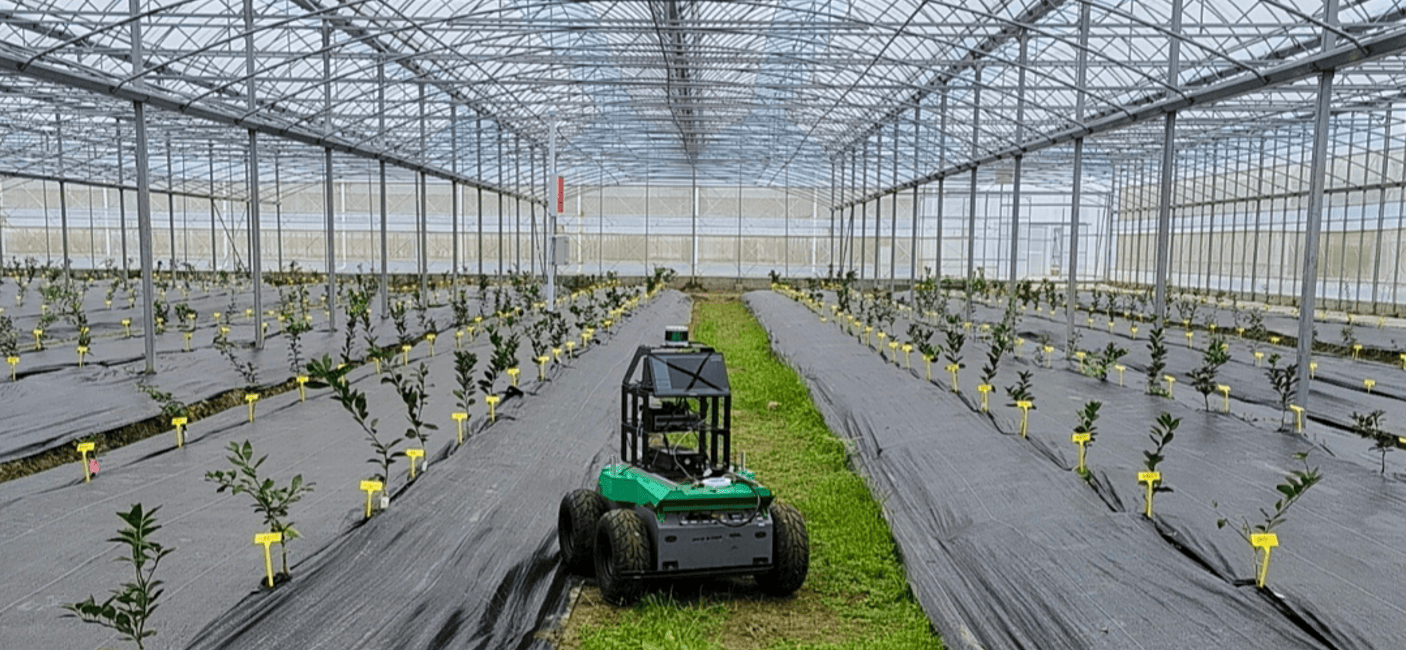}
    \caption{Pheno-Robot system operates in the greenhouse.}
    \label{fig: demo}
\end{figure}

\begin{figure*}[h]
    \centering
    \includegraphics[width=1.0\linewidth]{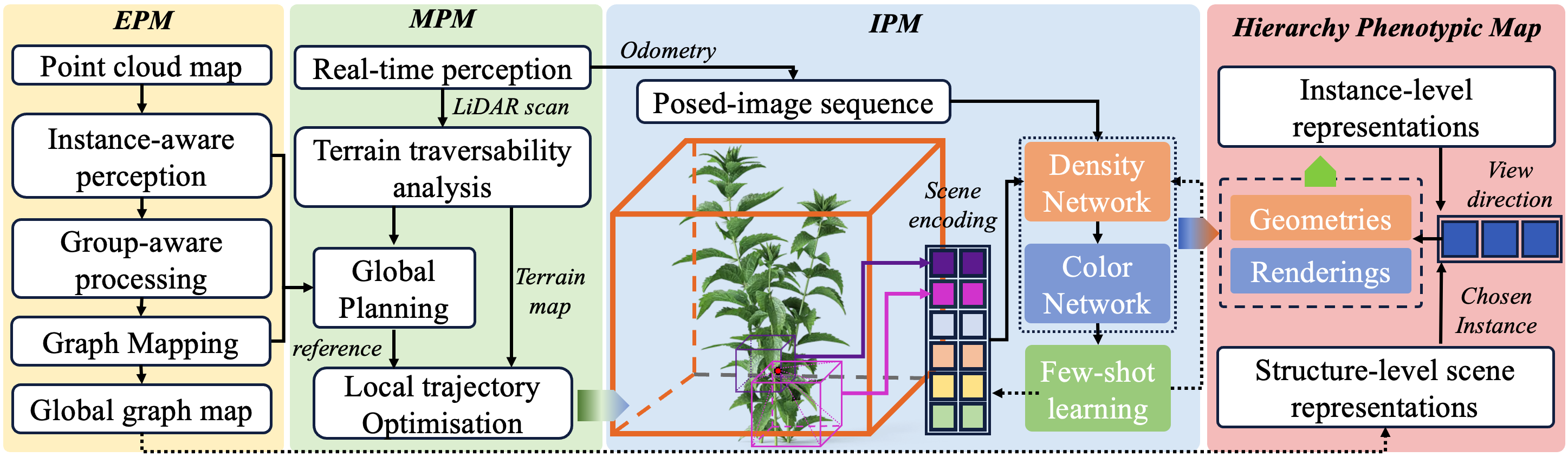}
    \caption{System Overview of the Pheno-Robot system.}
    \label{fig: overview}
\end{figure*}

Despite significant advances, the current practice heavily relies on manual operation, which is highly labour-intensive and inefficient, making it impractical for large-scale farms. 
Robotics offers the potential for widespread sampling in the field under authentic agricultural conditions. 
However, the realisation of robotic automation for high-quality in-situ plant phenotyping in the field faces two main challenges.
Firstly, there is a lack of effective methods for perception and autonomously navigating the intricacies of farm landscapes to perform large-scale and direct phenotyping. 
Secondly, there is a crucial gap in the availability of an accurate and effective digital modelling method capable of generating high-fidelity and multi-modal plant models. 
Successfully overcoming these challenges holds the key to enabling capabilities for repeated and detailed assessments of plants, potentially rendering a paradigm shift in the development of agri-genetics.

In this research, we present an innovative robotic system designed for autonomous in-situ phenotyping in the field, synthesising both robotic and digital technologies. 
The developed Pheno-Robot system comprises a comprehensive framework that includes a deep learning model and a motion planning method for robotic automation. 
It also incorporates a Neural Radiance Field (NeRF)-based modelling network, which enhances the system's ability to perform detailed and accurate in-situ phenotyping.
Our key contributions are:
\begin{itemize}
    \item develops a novel environmental understanding and the robotic navigation method tailored for farm environments.
    \item develops an improved NeRF method to address sparse-view input to achieve good quality in plant modelling.
    \item presents a hierarchy mapping method and demonstrates the Pheno-Robot system in agricultural settings. 
\end{itemize}

The rest of this paper is organised as follows.  Section \ref{section: review} surveys related work, followed by the proposed methodologies in Section \ref{section: method}. The experiment results are given in Sections \ref{section: experiment}. and then the conclusions are given in Section \ref{section: conclusion}.

\section{Related Works} \label{section: review}
\subsection{Robotic Automation in Agriculture}
Robots play a key role in precision agriculture, using smart and intervention technologies to improve efficiency through sensing and automation. 
Precision agriculture addresses spatial and temporal variability in the environment and plant growth patterns, scaling from traditional farm level \cite{virlet2016field} to sub-field precision \cite{mahmud2020robotics}. 
Horticulture farms are more complex than crop farms, requiring robots to navigate semi-structured, challenging terrain in dynamic environments \cite{duckett2018agricultural}. 
Firstly, sensor information is essential for detecting objects and potential risks in the field to ensure the safe operation of robotic vehicles. 
Secondly, machine learning is indispensable for farming robots to maximize locomotion flexibility (e.g., moving sideways or navigating narrow spaces between crops) and optimize input utilization with greater efficacy. 
The agility of these robots, coupled with the ability to carry specialized sensors, holds the key to unlocking the full potential of precision agriculture.
Despite notable progress, exemplified by robots designed for autonomous operations in orchard environments for selective harvesting \cite{au2023monash} or monitoring applications \cite{albani2017monitoring}, there is still uncharted territory. 
Robots capable of precise environmental recognition, robust and flexible motion, and accurate plant modelling remain an area for exploration.


\subsection{In-situ phenotyping}
Phenomics encompasses the study of various phenotypic plant traits, including growth, yield, plant height, leaf area index and more. 
Traditional methods rely on imaging and 3D range sensors to measure various plant traits such as colour, shape, volume, and spatial structure \cite{paulus2019measuring}. 
For example, the growth rate of rosette plants such as Arabidopsis \cite{jansen2009simultaneous} and tobacco \cite{dellen2015growth} is often analysed using images taken from a single viewpoint.
On the other hand, 3D range sensors allow accurate measurement of plant geometry and characteristics such as height, width and volume \cite{duckett2018agricultural}. 
Kang et al. \cite{kang2020real} proposed a sensor fusion system for yield estimation in apple orchards, using deep learning-based panoptic segmentation algorithms. 
Wu et al. \cite{wu2020mvs} utilised Multi-View Stereo (MVS) for the reconstruction of crop geometry in the field. 
However, 3D range sensors face limitations in agricultural environments due to strong occlusion, often resulting in a bleeding effect \cite{kang2022accurate} near discontinuous geometry. 
In addition, their point resolution is sparse compared to image-based data.
While MVS-based methods offer better quality with sufficient data, they require hours to process even simple plants \cite{xiong2016leaf}, limiting their applicability in real-time operations. 
Furthermore, current technologies do not provide a multi-modal representation of instances, making them inefficient for analysing plant features when multiple data types (image/geometry) are required.

\section{Methods} \label{section: method}
The Pheno-Robot system comprises three subsystems: an Environmental Perception Module (EPM), a Motion Planning Module (MPM), and an In-situ Phenotyping Module (IPM), as illustrated in Figure \ref{fig: overview}.

\subsection{Environment Perception}
\subsubsection{3D Detection Network}
\begin{figure}[h]
    \centering
    \includegraphics[width=0.8 \linewidth]{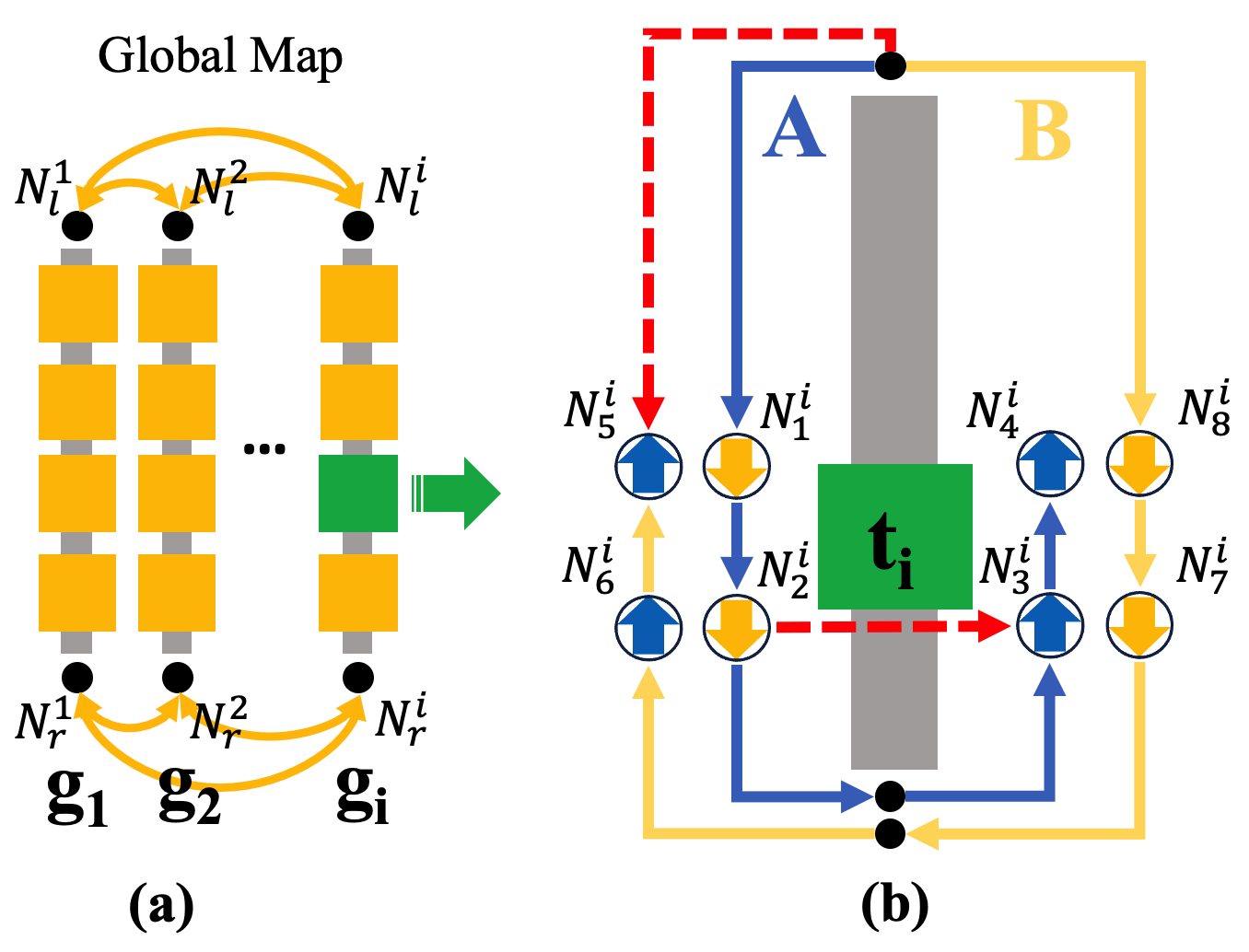}
    \caption{Illustration of EPM principle.}
    \label{fig: EPM}
\end{figure}

In this study, we utilise a novel neural network, the 3D Object Detection Network (3D-ODN), as introduced in our previous study \cite{pan2023novel}. 
Specifically designed for processing point cloud maps in Bird's-Eye View (BEV), the 3D-ODN consists of three integral network branches: point cloud subdivision, feature coding, and a detection head.
First, the entire point cloud map is uniformly segmentated. The points within each segment are then projected onto the BEV perspective, where local semantic features are extracted. 
These features are then fed into the main branch, forming a single-stage network architecture dedicated to processing the point cloud features from the BEV angle.
To improve the learning of multi-scale feature embeddings, the feature pyramid is used to facilitate the fusion of features of multi-scales. 
Finally, the recognition results are projected into 3D space based on the predicted height values. 
The identified instances are denoted as $\mathbf{T} = \{ t_{1}, t_{2}, \cdots \rvert t_{i} \in \mathbb{R}^2 \}$, representing the instances detected by the 3D-ODN (see Fig.\ref{fig: EPM} (a)).

\subsubsection{Graph Mapping}
For each instance, eight nodes are designated to represent the four corners in two different directions of motion, as shown in Fig.\ref{fig: EPM} (b). 
This orientation consideration aims to mitigate turning movements in narrow passages, which could otherwise lead to a higher failure rate of movement due to trapping into potential obstacles. 
Consequently, each instance $o$ is associated with eight nodes, denoted as $t_{i} = \{ N^{i}_{1}, N^{i}_{2}, \cdots, N^{i}_{8} \rvert N^{i}_j \in \mathbb{R}^2 \}$ (see Fig.\ref{fig: EPM} (b)).
To extract the line structure of farms, a line detection algorithm is used to group instances in the same row, denoted $\mathbf{g}$. 
Each row has two common access nodes at both ends, denoted $N_{l}$ and $N_{r}$ ($N_{l}$ and $N_{r} \in \mathbb{R}^2$). 
These nodes serve to establish connections between all instances within the group and to form links with other groups (see Fig.\ref{fig: EPM} (a)). A group is thus characterised as $g_{i} =\{ t_{1}, t_{2}, \cdots , N^{i}_{l}, N^{i}_{r} \}$. The map is then represented as a set of groups denoted by $\mathbf{G}=\{ g_{1}, g_{2}, \cdots \}$.

\subsection{Trajectory Planning}
\begin{figure}[h]
    \centering
    \includegraphics[width=0.8 \linewidth]{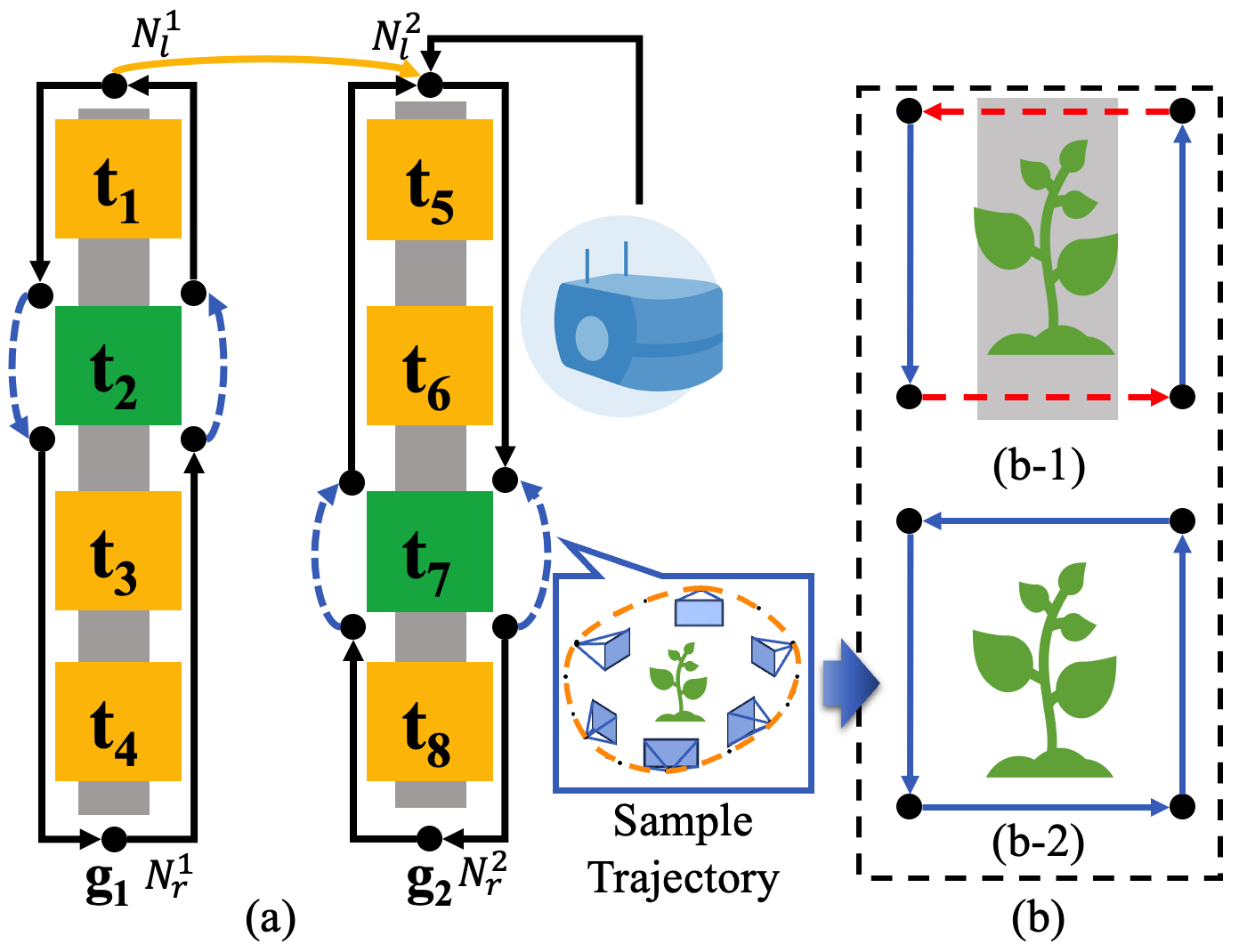}
    \caption{Illustration of MPM principle.}
    \label{fig: MPM}
\end{figure}

\subsubsection{Global Path Planning}
A Greedy-search-based path generation algorithm on the graph map is developed, as illustrated in Alg.1. The following are the relevant definitions:
\begin{itemize}
    \item $\mathbf{T}$ represents the goal set that requires phenotyping.
    \item $\mathbf{V}$ represents the subgroups if their instances in $\mathbf{T}$.
    \item $\mathbf{N_{start}}$ represents the robot position at the start.
    \item $\mathbf{\Gamma}$ represents the global path that includes a sequence of nodes.
\end{itemize}

\begin{algorithm}
\caption{Global Path Generation}
\begin{algorithmic}[1]
\REQUIRE $\mathbf{T}, \mathbf{G}, \mathbf{N_{start}}$   
\ENSURE $\mathbf{\Gamma}$   
\FOR {$t_{i}$ in $\mathbf{T}$}    
    \STATE $\mathbf{v}_{new} \leftarrow \mathbf{FindParent}(t_{i});$\\
    \STATE $\mathbf{V} \leftarrow \mathbf{V} \bigcup \mathbf{v}_{new};$
\ENDFOR
\WHILE{$\mathbf{V} \neq \mathbf{\emptyset}$}
    \STATE $\mathbf{v}_{j} \leftarrow \mathbf{FindNearestSubgroup}(\mathbf{\Gamma}, \mathbf{V});$
    \STATE $\mathbf{\Gamma}_{new} \leftarrow \mathbf{PlanConnection}(\mathbf{\Gamma}, \mathbf{v}_{j});$
    \STATE $\mathbf{\Gamma} \leftarrow \mathbf{\Gamma} \bigcup \mathbf{\Gamma}_{new};$
    \STATE $\mathbf{Delete} \ \mathbf{v}_{j}$ $\mathbf{from} \ \mathbf{V};$
\ENDWHILE
\RETURN $\mathbf{\Gamma}$
\end{algorithmic}
\end{algorithm}

\noindent Two key sub-functions presented in Alg.1 are described as follows:
\begin{itemize}
    \item $\mathbf{FindParent}$: this function finds the group that this instance belongs to.
    \item $\mathbf{FindNearestSubgroup}$: this function finds the subgroup that has the instance closest to the current robot location.
\end{itemize}

\noindent The key functions $\mathbf{PlanConnection}$ is described in Alg.2.

\begin{algorithm}
\caption{$\mathbf{PlanConnection}$}
\begin{algorithmic}[1]
\REQUIRE $\mathbf{\Gamma}, \mathbf{v}=\{ t_{1}, t_{2}, \cdot , t_{n} \rvert t_{i} \in \mathbf{g}_{i}\}$   
\ENSURE $\mathbf{\Gamma}_{new}$   
\WHILE {$\mathbf{v} \neq \mathbf{\emptyset}$}    
    \STATE $\mathbf{N}_{new}, \ t_{i} \leftarrow \mathbf{FindNearestAndFeasible}(\mathbf{\mathbf{\Gamma}, v});$\\
    \STATE $\mathbf{\Gamma}_{new} \leftarrow \mathbf{\Gamma}_{new} \bigcup \mathbf{N}_{new};$
    \IF{$\mathbf{IFFullyCover}(\Gamma_{new}, t_{i})$}
    \STATE $\mathbf{Delete} \ t_{i}$ $\mathbf{from} \ \mathbf{v};$
    \ENDIF
\ENDWHILE
\RETURN $\mathbf{\Gamma}_{new}$
\end{algorithmic}
\end{algorithm}

\noindent  The function $\mathbf{FindNearestAndFeasible}$ finds the feasible and the nearest node (see Fig.\ref{fig: MPM} (a)) of an instance in the subgroup. It evaluates the feasibility from two perspectives: traversability and orientation, which will be detailed in Sec \ref{section: traversable_analysis}. Another function $\mathbf{IFFullyCover}$ evaluates if an instance has been fully covered by the generated path. This is achieved by checking the number of connected nodes of an instance, as shown in Fig.\ref{fig: MPM} (b).

\subsubsection{Local Trajectory Generation}
This step aims to compute the detailed path based on the global path, including three steps: path generation, optimisation, and interpolation.

\textbf{Initial path Generation}: 
Given the global path $\Gamma$, the first step is to generate an initial collision-free path between two nodes. 
Two scenarios are considered, including path generation for phenotyping data acquisition (where two adjacent nodes belong to the same instance) and paths between nodes of different instances.
In the first case, a sequence of collision-free preset sample positions is established and the RRT algorithm is used to connect these positions to form the sampling trajectory between nodes.
In the second case, the $\mathbb{A}^{*}$ algorithm is used to determine a collision-free path between nodes.

\textbf{Trajectory Optimisation}: 
The initial path $\Phi$ consists of a sequence of points $ \{ Q_{0}, Q_{1},\cdots | Q \in \mathbb{R}^{2} \}$, and we conceptualise a trajectory $t \in [0,1] \rightarrow \Phi \subset \mathbb{R}^2$ as a continuous function mapping time to robot states.
The objective function incorporates three key aspects of the robot's motion. 
It penalises velocities to encourage smoothness, proximity to the environment to secure trajectories that maintain a certain distance from obstacles, and the distance between the state and the preset viewpoint. 
These terms are represented as $f_{s}$, $f_{c}$, and $f_{o}$, respectively. The objective function is:
\begin{equation}
    \Phi^{*} = \mathop{\mathbf{argmin}}_{\Phi} \ \alpha_{s} f_{s}(\Phi) + \alpha_{c} f_{c}(\Phi) +\alpha_{o} f_{o}(\Phi)
\end{equation}

\noindent the initial state $\Phi(0)=Q_{0}$ and final state $\Phi(1)=Q_{1}$ are fixed. 
$\alpha_{s}$, $\alpha_{c}$, and $\alpha_{o}$ are weights for each penalty terms.
The detailed formulations of $f_{s}$, $f_{c}$, and $f_{o}$ are as follows:
\begin{equation}
    f_{s}(\Phi) = \int_{\Phi} || \frac{d}{dt}\Phi(t) ||^{2} 
\end{equation}
\begin{equation}
    f_{c}(\Phi) = \int_{\Phi} c(\Phi(t)) || \frac{d}{dt}\Phi(t) || dt
\end{equation}
\begin{equation}
    f_{o}(\Phi) = \int_{\Phi} || \Phi(t) - \Phi_{d} ||^{2} dt
\end{equation}
\noindent where $c(\cdot): \mathbb{R}^{2} \to \mathbb{R}$ be the function that penalises the state near the obstacles, $\Phi_{d}$ is the trajectory that contains the desired view-position for data acquisition.

We update the trajectory by the functional gradients $\bar{\nabla} f(\Phi_{i})$ using $\Phi_{i+1} = \Phi_{i}-l_{\mathbf{r}} \cdot \bar{\nabla} f(\Phi_{i})$, following the work \cite{zucker2013chomp}, where $l_{\mathbf{r}}$ is the learning rate. The functional gradient of the objective in (2)-(4) is given by
\begin{equation}
    \bar{\nabla} f_{s}(\Phi) = -\frac{d^{2}}{dt^{2}} \Phi(t) 
\end{equation}
\begin{equation}
    \bar{\nabla} f_{c}(\Phi) = ||\Phi'(t)|| \cdot [(I-\Phi'(t)\Phi'(t)^{T}) \nabla c - c \kappa]
\end{equation}
\begin{equation}
    \bar{\nabla} f_{o}(\Phi) = -(\Phi(t)-\Phi_d)
\end{equation}
where $\nabla c$ is the derivative of obstacles to the control points $Q_{j}$ by the $\partial c / \partial Q_{j}$, the definition of $\kappa$ is given as below.
\begin{equation}
    \kappa = ||\Phi'(t)||^{2}((I-\Phi'(t)\Phi'(t)^{T})\Phi'')
\end{equation}

\textbf{B-spline Interpolation}:
The optimised trajectory is then parameterised by a piece-wise B-spline into a uniform curve $\Phi_{B}$. 
Given the determined degree $m$, and a knot vector $\{ k_{0}, k_{1}, \cdot, k_{M} \}$, where $M = n_{q}+ 2m$. The parameterised uniform curve $\zeta(t)$ can be formulated as:
\begin{equation}
    \zeta(t) = \frac{\sum\limits_{i=1}^{n_{q}}R_{i,m}(t)w_{i}Q_{i}}{\sum\limits_{i=1}^{n_{q}}R_{i,m}(t)w_{i}}
\end{equation}
where $w_{i}$ is the weight of each control points $Q_{i}$, $R_{i,m}$ is the basis function for $Q_{i}$ at $m$ degree, $u$ is the control value of the curve.
Then, the TEB-planner \cite{rosmann2017kinodynamic} is used to find the proper velocity to follow the trajectory.

\subsubsection{Feasibility checking} \label{section: traversable_analysis}
The $\mathbf{FindNearestAndFeasible}$ function in Alg.2 evaluates the feasibility between two nodes from two aspects: \textbf{orientations} and \textbf{traversability}.

\textbf{I. Orientation Checking}: This term evaluates whether two nodes have similar directions, as the narrow tunnels between plants do not allow turn-round. The largest allowed orientation difference between two nodes is $1 / 3\pi$.

\begin{figure}[ht]
    \centering
    \includegraphics[width=1\linewidth]{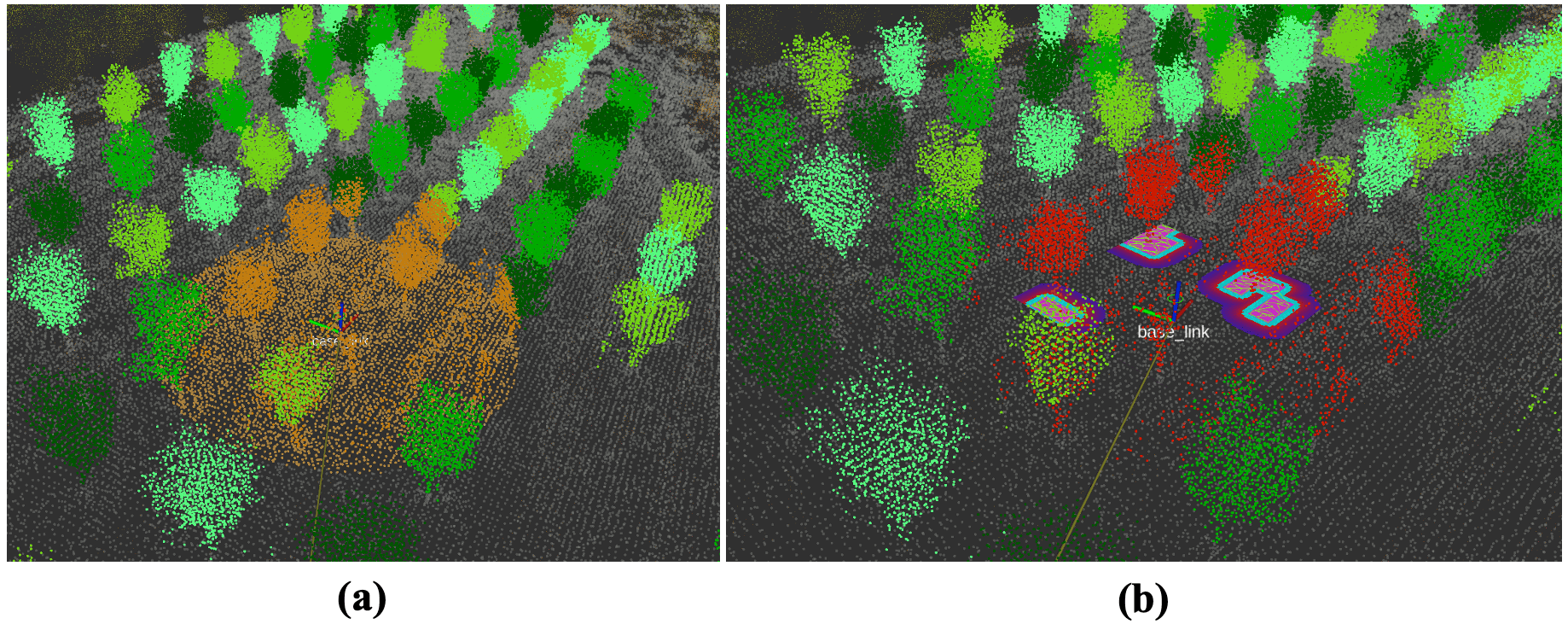}
    \caption{Illustration of terrain analysis, a) the points of robot surroundings, b) the risk area is converted into polygons.}
    \label{fig:local terrain}
\end{figure}

\textbf{II. Traversability Analysis}: 
The point cloud between two nodes is projected onto a 2D grid map. 
The traversability of each grid is assessed as the multiple risk factors, including:

\textbf{\textit{(a) Collision risk}}: A risk factor quantified by the possibility of a grid belonging to obstacles, denoted $\theta \in [0,1]$. 
A terrain analysis \cite{CSF} is utilised here.

\textbf{\textit{(b) Slope risk}}: For each point on the terrain $T_{i}$, points on the map are selected by a cube box with sides of length $l_{s}$, which is represented as $\Omega_{i} = \{ (p^{j}_{i})_{j=1:N_{i}} \rvert p_{i}^{j} \in \mathbb{R}^{3}\}$. The SVD is used to fit a plane $P_{i}$ from the $\Omega_{i}$ and get the normal vector $n_{z} \in \mathbb{R}^{3}$. The slop angle between terrain $T_{i}$ and vertical direction $n_{z}$ is obtained by $\mathbf{s} = \mathbf{arccos}\frac{||e_{z} \cdot n_{z}||}{||e_{z}|| \cdot ||n_{z}||}$.

\textbf{\textit{(c) Step risk}}: This term evaluates the height gap between adjacent grid cells. Negative obstacles can also be detected by checking the lack of measurement points in a cell. The maximum height gap is denoted as $\lambda$.

The above three risks are combined by a weighted sum to obtain the weighted traversability value $\Upsilon$, as:
\begin{equation}
    \Upsilon = \theta + \alpha_{\mathbf{s}} \frac{\mathbf{s}}{\mathbf{s}_{crit}}+\alpha_{\lambda} \frac{\lambda}{\lambda_{crit}}
\end{equation}
where the $\mathbf{s}_{crit}$ and $\lambda_{crit}$ are the maximum allowed slope angle and height gap, respectively.

\textbf{III. Terrain analysis for trajectory optimisation}: 
For robot operation, the odometry and KD-tree \cite{cai2021ikd} are utilised to build the local terrain map (Fig.\ref{fig:local terrain} (a)). 
The detected obstacle points (Fig.\ref{fig:local terrain} (b)) are converted to polygons, denoted as $\{\mathcal{O} \subset \mathcal{R}^{2}\}$. 
Let the $\phi(Q_{i}, \mathcal{O})$ describe the minimal Euclidean distance between obstacles and a control point. 
A minimum separation $\phi_{\mathrm{min}}$ between all obstacles and $Q_{i}$ is found by
\begin{equation}
    \phi_{\mathrm{min}} = \mathrm{min}[\phi(Q_{i}, \mathcal{O}_{1}),\phi(Q_{i}, \mathcal{O}_{2}),\cdots]
\end{equation}
The obstacles $\mathcal{O}$ is updated online for dynamic environments.

\subsection{In-situ Phenotyping Model}
\subsubsection{Neural Rendering}
NeRF represent the 3D scene as a radiance field which describes volume density $\sigma$ and view-dependent color $c$ for every point $x$ and every viewing direction $d$ via a MLP \cite{mildenhall2021nerf}:
\begin{equation}
    \label{EQ:NERFMLP}
    \sigma,c=H_{\Theta}(x,d)
\end{equation}
The ray tracing-based volume rendering is used to render the parameters of a 3D scene into colours $\hat{C}$ of a ray $r$, expressed as:
\begin{equation}
    \label{Eq:volume rendering}
   \begin{aligned}\hat{C}(r)&=\sum_{i=1}^NT_i\left(1-\exp\left(-\sigma_i\delta_i\right)\right)c_i,T_i\end{aligned}
\end{equation}
Where $T$ is the volume transmittance and $\delta$ is the step size of ray marching. 
For every pixel, the squared loss on photometric error is used for the optimization of MLP. 
When applied across the image, this loss $\mathcal{L}_{\mathrm{Rendering}}$ is represented as: 
\begin{equation}
    \label{Eq:L}
   \mathcal{L}_{\mathrm{Rendering}}=\sum_{r\in R}||\hat{C}(r)-C(r)||_2^2
\end{equation}
where $C(r)$ is the ground truth colours. 

\begin{figure}[ht]
    \centering
    \includegraphics[width=1\linewidth]{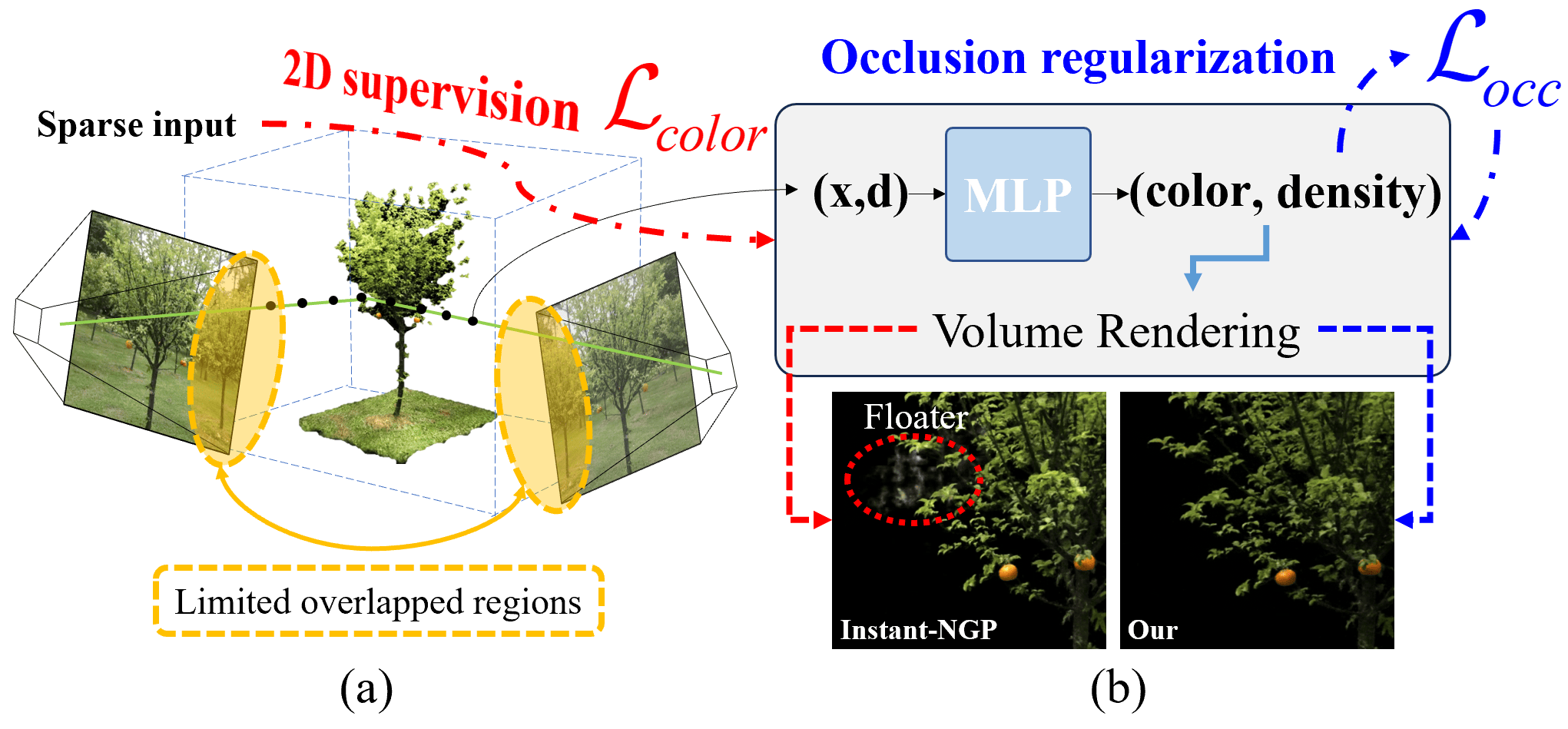}
    \caption{Few-shot learning from sparse input. a) Ray samples under limited training views. b) Occlusion regularization for sparse-view rendering.}
    \label{fig:few-shot}
\end{figure}

\subsubsection{Few-shot learning from Sparse Views}  
The artefact of "white floaters" caused by rendering distortion is the common failure mode in NeRF learning.
Autonomous data acquisition by robots can always lead to imperfect density and sparse views with fewer overlapping regions (Fig. \ref{fig:few-shot} (a)), thus distorting the rendering of these regions. 
This is essential because, between these sparse views, NeRF's training lacks sufficient information to estimate the correct geometric information of the scene, leading to significant and dense floats in the region that is close to the camera. 
To reduce those artefacts, an optimisation term $\mathcal{L}_{occ}$, which is designed to regulate the learning behaviour of NeRF \cite{yang2023freenerf}, is used to penalize the dense fields near the camera via “occlusion” regularization, which can be expressed as:
\begin{equation}
    \label{EQ:occ}
    \begin{aligned}
        \mathcal{L}_{occ}&=\frac{\boldsymbol{\sigma}_{K} \mathbf{m}_{K}}{K}=\frac{1}{K}\sum_{K}\sigma_k\cdot m_k
    \end{aligned}
\end{equation}
where $\boldsymbol{\sigma_K}$ represents the density values of the $K$ points sampled along the ray, ranked in order of proximity to the origin, $\mathbf{m_K}$ is the binary mask vector that determines whether a ray sector will be penalised or not. 

\subsubsection{Geometry Extraction from Field} \label{section:nerf2mesh} 
Given a predefined 3D region of interest, a set of spatial points $P=\{p_1,p_2,...,p_n\}$ is generated via dense volumetric sampling. For each point $p_i\in P$, it's evaluated through the NeRF model to obtain The density values, $\sigma(p_i)=\mathrm{NeRF}_\sigma(p_i)$, form the basis for surface extraction.
The Marching Cubes algorithm identifies the iso-surface by threshold of the density values: 
\begin{equation}
    M=\mathbf{MarchingCubes}(P,\sigma_{\text{threshold}})
\end{equation}
Where $M$ is the resultant mesh and $\sigma_{threshold}$ is an optimal density value demarcating the object's boundary.  

For vertex $v_j$ in mesh $M$, a viewing ray $r_j$ is constructed and queried $ \mathbf{c}(v_j)=\mathrm{NeRF}_\mathbf{c}(v_j,r_j)$ in NeRF. 
Those values derived from the field are mapped onto mesh $M$, assigning colour to each vertex: 
\begin{equation}
    \operatorname{Color}(v_j)=\mathbf{c}(v_j)
\end{equation}
For a 2D texture representation, the vertex-coloured mesh undergoes UV unwrapping. 
To minimize distortion, Least Squares Conformal Mapping (LSCM) \cite{levy2023least} is used, which is to minimize the conformal energy: 
\begin{equation}
    E(u,v)=\int_{\Omega}\left(|\nabla u|^2+|\nabla v|^2\right)dA
\end{equation}
where $(u,v)$ are the 2D texture coordinates for each vertex in $M$, $\Omega$ represents the object's surface, and $dA$ is a differential area element on the mesh's surface, indicating that the energy is computed by integrating over the surface of the mesh.

\subsubsection{NeRF Model} 
The Instant-NGP \cite{muller2022instant}, which makes use of a multi-resolution hashed encoding that can represent learned features of the scenes with tiny MLPs, is utilised. 
In detail, Instant-NGP operates on the premise that the object to be reconstructed is enclosed within multi-resolution grids.  
For any point $\mathbf{x}\in\mathbb{R}^{3}$ in various resolution grids, it obtains the hash encoding $h^i(\mathbf{x})\in\mathbb{R}^d$, where $d$ is the features' dimension, $i$ is the level of tri-linear interpolation. 
The hash encodings of all levels are concatenated to form the multi-resolution feature $h(\mathbf{x})=\{h^i(\mathbf{x})\}_{i=1}^L\in \mathbb{R}^{L\times d}$.

\textbf{Training}: The primary goal of our study is to produce high-quality rendering models of instances. 
We utilise the rendering loss as it quantifies the discrepancy between the rendered images and the input images, denoted as ${\mathcal{L}_{color}}$. 
Besides, to improve the MLP learning under sparse view, we also apply ${\mathcal{L}_{occ}}$ from (12).
Therefore, given a set of posed-images \(I_{gt}\) and the predicted renderings from the network \(I_{pred}\), the training loss ${\mathcal{L}_{MLP}} $ is defined as:
\begin{equation}
{\mathcal{L}_{MLP}} = {\mathcal{L}_{color}}+ \mathcal{L}_{occ}
\end{equation}
\noindent where ${\mathcal{L}_{color}} = (1/N) \cdot \sum_{i=1}^{N} \| I_{gt}^{(i)} - I_{pred}^{(i)} \|_2^2$ and $N$ is the total number of images in the datasets.

\subsection{Hierarchy Map Representation}
A hierarchy scene representation that combines large-scale mapping with local high-fidelity rendering is introduced, including structure-level and instance-level.

\textbf{I. Structure-level representation}: The coarse-level map is designed to extract high-level information, focusing on the structure of the scene and the plants in the environment. 
In this context, we construct a map of the farm by creating a coloured point cloud map and its corresponding semantic map.

\textbf{II. Instance-level representation}: Each plant in the environment will be meticulously represented, ensuring that each plant $o_{i}$ is captured in fine detail through the robot's visual image stream. In the context of phenotyping, each detected plant will be associated with a high-fidelity neural-learned rendering model, accompanied by its corresponding 3D model.

\section{Experiments} \label{section: experiment}
\subsection{Pheno-Robot Hardware}
\begin{figure}[ht]
    \centering
    \includegraphics[width= 0.9\linewidth]{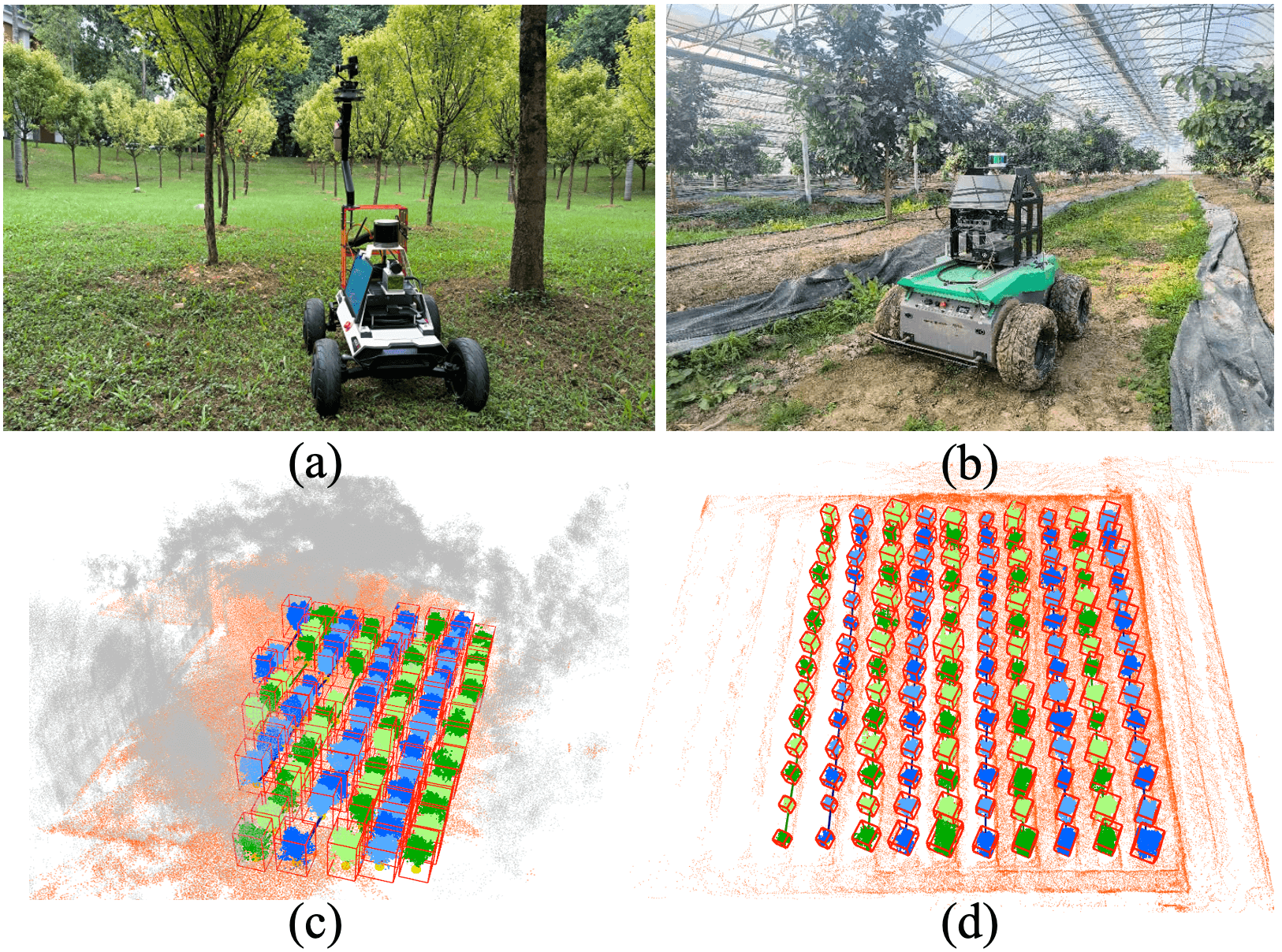}
    \caption{(a) and (b): Pheno-Robot systems; (c) and (d): the demonstration of the EPM results in two test scenarios.}
    \label{fig: epm_result}
\end{figure}

The Pheno-Robot is equipped with a 32-line LiDAR, a Realsense D435 camera and a 9-axis IMU (see Fig.\ref{fig: epm_result} (a) and (b)) for navigation.
Communication between the different modules is via the common message layer on the ROS server.
The sensors are connected to the computer via ROS-Noetic on Ubuntu 20.04.
A 4K GoPro Hero-11 is mounted on a gimbal stabilizer on the left side, and the image stream from the GoPro is transmitted to the computer through the GoPro-ROS-node.
Plant modelling by NeRF training is performed remotely, with data transmission via the 5G wireless network.

\subsection{Evaluation on Pheno-Robot system}

\subsubsection{Evaluation on EPM}
This section evaluates the performance of the EPM in typical agricultural environments. 
Two different settings, shown in Fig. \ref{fig: epm_result} (a) and (b), were selected for the system evaluation, and their corresponding semantic maps are shown in Fig. \ref{fig: epm_result}(c) and (d), respectively. 
The results show the precise instance-level understanding of the EPM of the point cloud map in agricultural scenarios, with an overall recall and precision for detection of 0.95 and 1.0, respectively. 
The average position error and bounding box error are measured to be 0.06m and 0.02m respectively.
Particularly in agricultural landscapes with row structures, our method demonstrates the ability to detect such features and generate a graph map for robot autonomy. Furthermore, the method is versatile and supports both online and offline modes depending on specific requirements. For this study, we perform the semantic extraction process offline.

\subsubsection{Evaluation on MPM}
\begin{figure}[ht]
    \centering
    \includegraphics[width=0.9\linewidth]{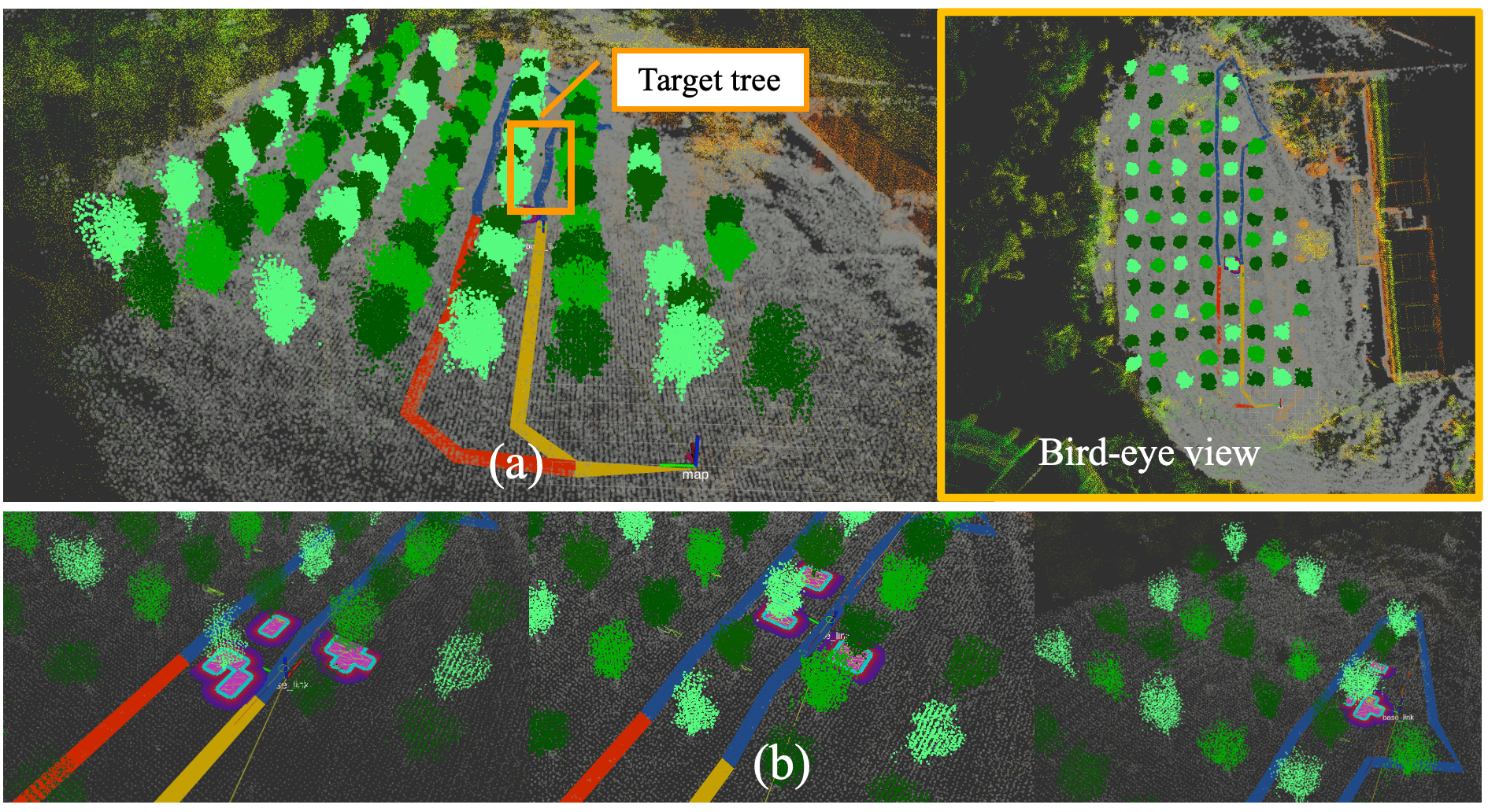}
    \caption{Planning scene and motion planning results of the MPM in orchard-like environments.}
    \label{fig: mpm_result1}
\end{figure}

\begin{figure}[ht]
    \centering
    \includegraphics[width=0.9\linewidth]{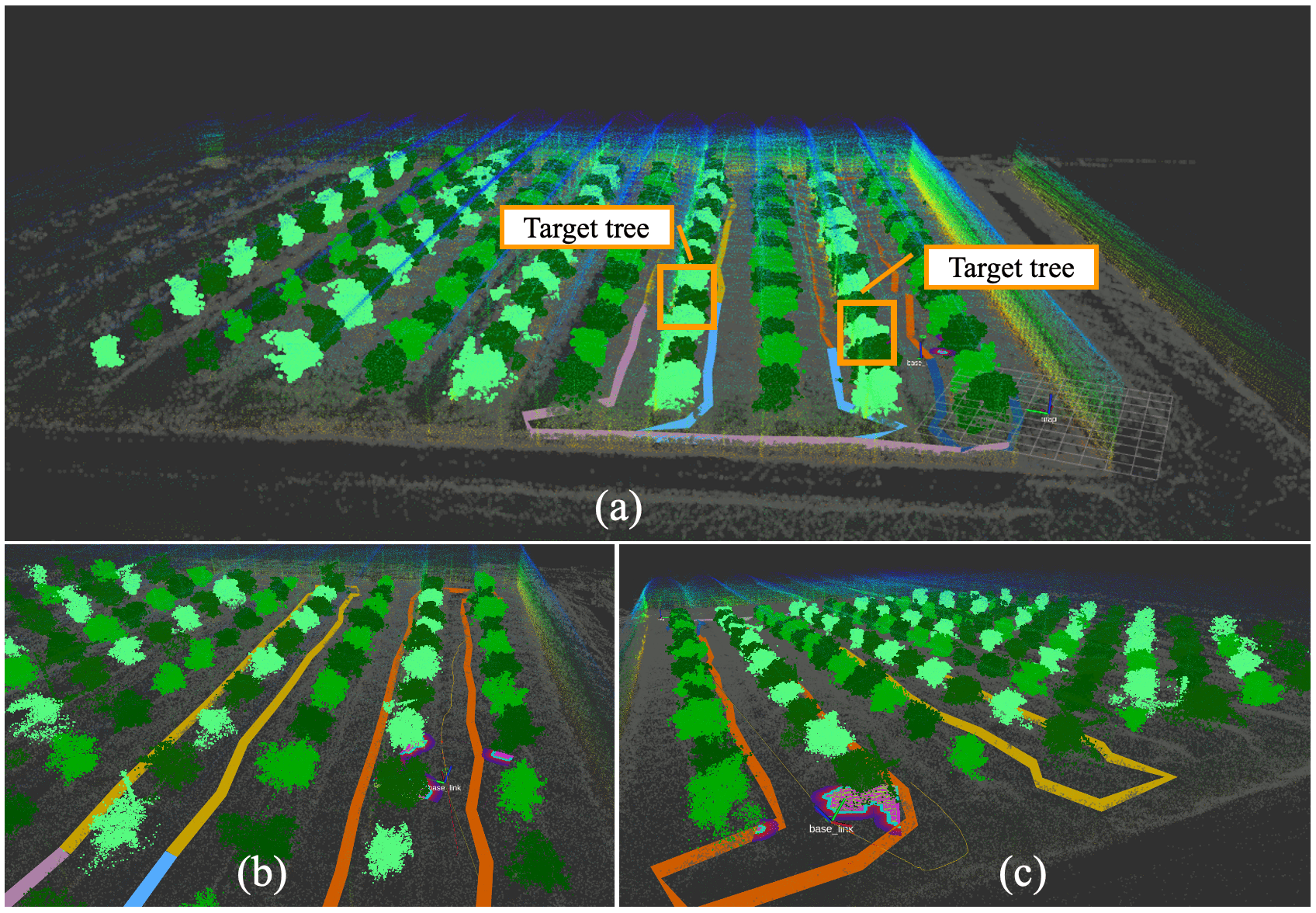}
    \caption{Planning scene and motion planning results of the MPM in greenhouses.}
    \label{fig: mpm_result2}
\end{figure}

This section evaluates the performance of the MPM for automated phenotyping by robots. 
We conducted tests in two different environments, shown in Fig. \ref{fig: mpm_result1} and \ref{fig: mpm_result2}, corresponding to the environments shown in Fig. \ref{fig: epm_result} (a) and (b), respectively.

In the experiment, target plants requiring phenotyping were randomly selected. 
The results show that the presented MPM achieves a high success rate and meets the robot's requirements in both scenarios, as shown in (a) of Fig. \ref{fig: mpm_result1} and \ref{fig: mpm_result2}. 
Compared to conventional $A^{*}$ or Dijkstra planners, which may generate inappropriate global trajectories leading to turning movements in narrow channels, causing local motion planning failures for car-like robots, our graph-based global planner performs better.
Beyond the global planner, the developed local trajectory planner also shows strong robustness in field environments, as illustrated in (b) and (c) of Figs. \ref{fig: mpm_result1} and \ref{fig: mpm_result2}, respectively. 
A significant factor affecting the performance of the local trajectory planner is the traversability analysis. 
Given the complex terrain in agricultural environments, traversability may be prone to overestimating or underestimating risk areas during planning, leading to motion planning failures. 
In the experiment, the robot was optimised with an additional solid-state lidar at the front to improve its perception under complex conditions. 
In addition, a replanning mechanism is employed for local motion planning, with an update frequency of 5HZ and a forward planning distance of 10m.
Overall, our system demonstrates robust performance in the field with appropriate parameter tuning. 
The maximum speed during sampling is 0.2m/s, while the maximum speed for other conditions is 1m/s.

\begin{figure*}[ht]
    \centering
    \includegraphics[width=0.92\linewidth]{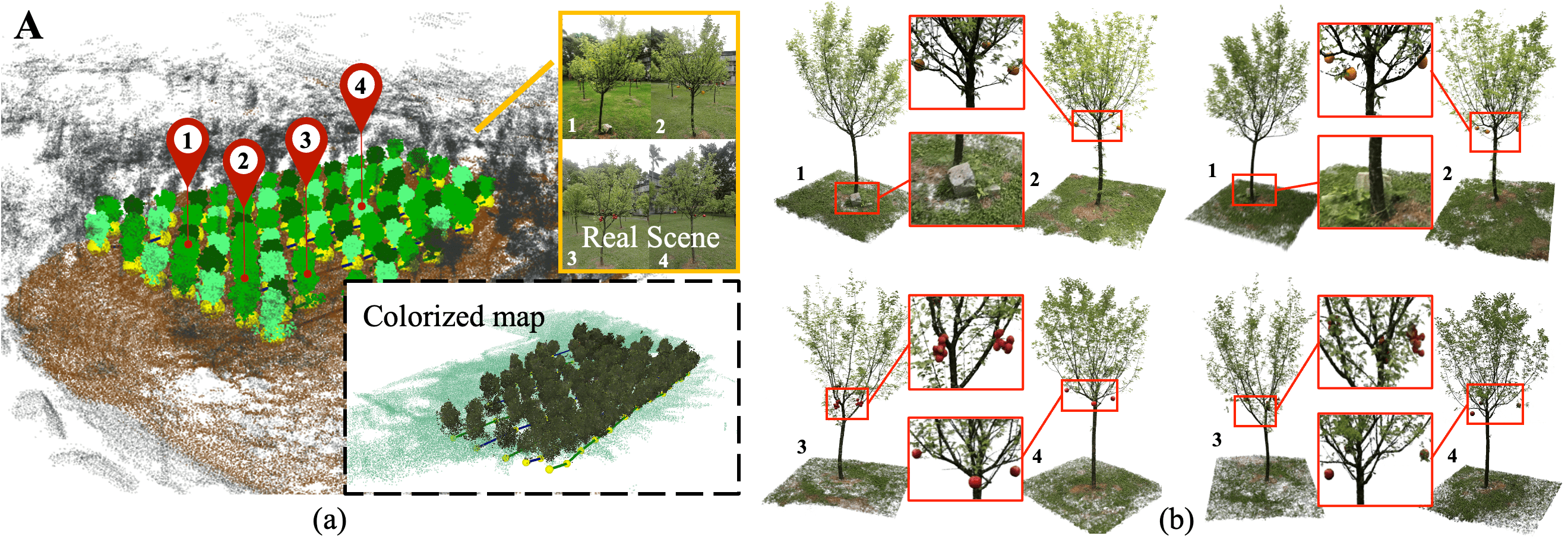}
    \includegraphics[width=0.92\linewidth]{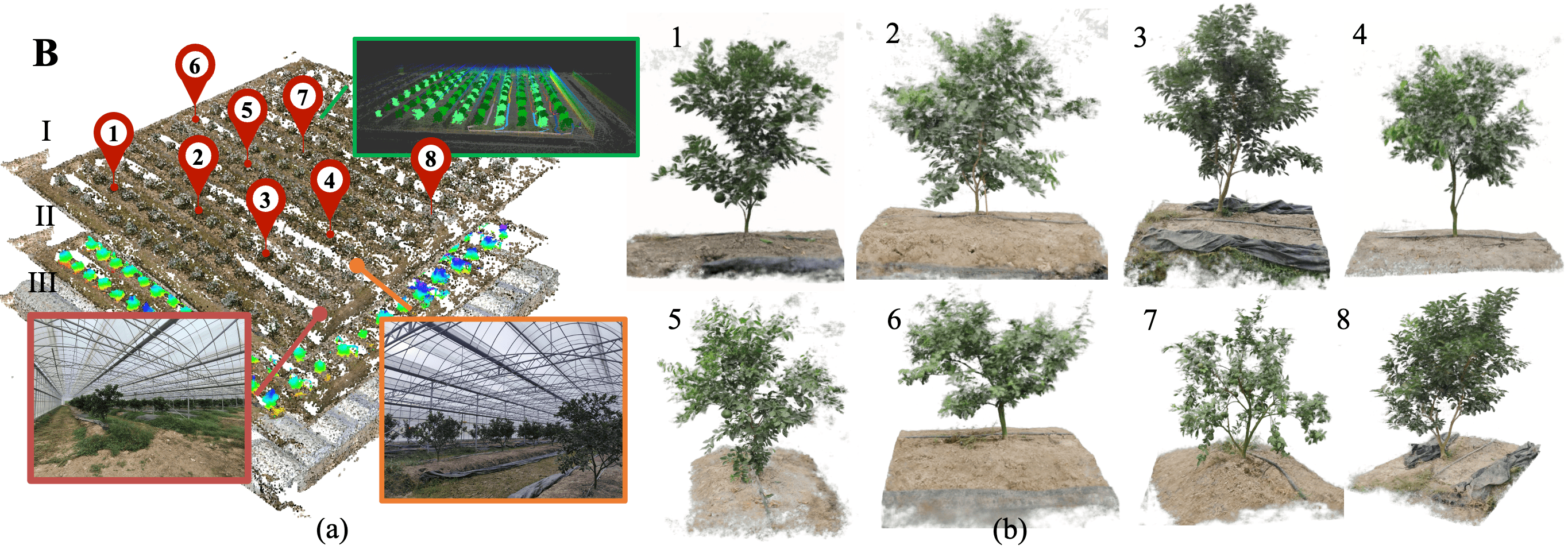}
    \caption{Demonstration of hierarchy maps of two scenarios in orchard-like environments (A) and greenhouse (B). The a) of both A and B are global-level maps of the scenes, and b) are the detailed models of instances in the environments.}
    \label{fig:enter-label-1}
\end{figure*}

\subsubsection{Evaluation on in-situ Phenotype}
This section assesses the performance of IPM. 
We compare the in-situ phenotyping models of both scenarios by using handheld data acquisition and robot-automated acquisition. 
The quality of the results is evaluated by Peak Signal to Noise Ratio (PSNR), as detailed in Table 1.

\begin{table}[ht]
\centering
\scriptsize
\caption{Comparison of Modelling quality}
\begin{tabular}{l|l|c| c|c| c}
\Xhline{1pt}
& & \multicolumn{2}{c|}{\textbf{PSNR(dB)}} & \multicolumn{2}{c}{\textbf{Time(min)}} \\
\cline{3-6}
 \multirow{-2}{*}{\textbf{Dataset}} &  \multirow{-2}{*}{\textbf{Methods}} & \textbf{HA} & \textbf{RA} & \textbf{HA} & \textbf{RA} \\
\Xhline{1pt}
\multirow{2}{*}{\textbf{Outdoor}}  & Instant-NGP 
& 24.8 & 22.5 & 2.4  & 3.2  \\  
\cline{2-6}
& \textbf{Ours} & 25.2& 24.2 & 2.3  & 2.7   \\
\cline{1-6}
\multirow{2}{*}{\textbf{Greenhouse}}  
& Instant-NGP  & 23.4   & 21.5 & 3.1 & 5.2   \\  
\cline{2-6}
& \textbf{Ours} & 23.3   & 22.7 & 2.3  & 4.1   \\
\Xhline{1pt}
\end{tabular}
\label{tab:comparison_table}
\end{table}

In the initial experiments, both vanilla Instant-NGP and our model showed fast convergence, taking less than 2 minutes when using hand-collected samples, resulting in a PSNR above 24 dB and indicating high-quality results. 
However, when using robot-collected samples, Instant-NGP took over 4 minutes to converge and achieved a PSNR below 23 dB. 
In contrast, our method achieved a PSNR above 23.5 dB and converged in 3 minutes, demonstrating the effectiveness of occlusion regularisation in NeRF training with sparse-view inputs, effectively mitigating geometric estimation errors in the renderings.
Moving on to the greenhouse experiments, the quality of modelling using both Instant-NGP and our model showed a decrease, with PSNR values of around 23 dB in approximately 3 minutes when using handheld collected samples. 
This reduction in quality is mainly due to the complex geometry of the canopy leading to occlusion. 
When modelling with robotically collected samples, both models showed a decrease in model quality along with a longer training convergence time. 
In comparison, our model performed better under these field conditions.
The hierarchy maps for both scenarios are shown in Figures \ref{fig:enter-label-1} and \ref{fig:enter-label-2}, revealing global structure-aware maps of farms and detailed models for individual plants. 
These results highlight the ability of the robotic system to achieve high-quality in-situ phenotyping in complex agricultural environments, demonstrating resilience despite certain limitations.

\subsection{Demonstration of Digital-Modelling for Simulation}
\begin{figure}[ht]
    \centering
    \includegraphics[width=\linewidth]{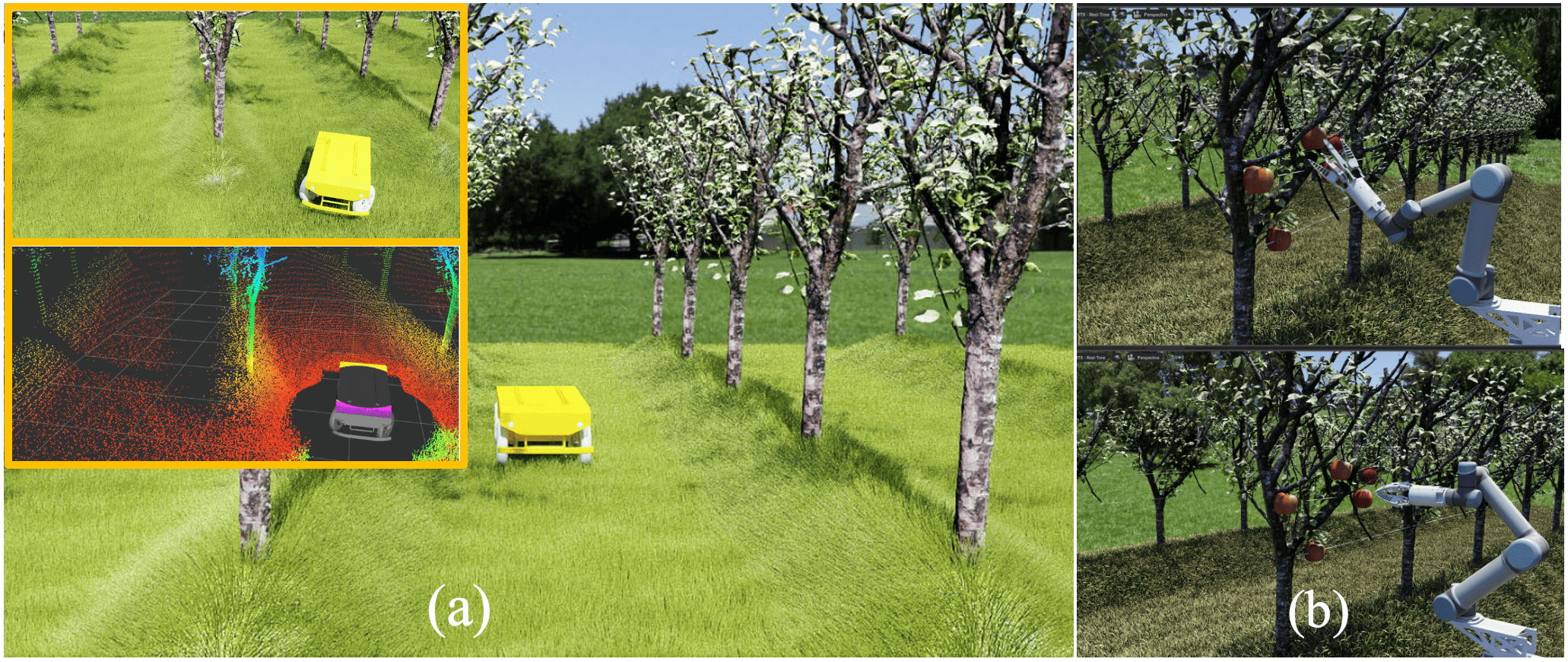}
    \caption{Isaac simulation environments of a) robot navigation and b) robot harvesting.}
    \label{fig: isaac_1}
\end{figure}

Our system can streamline the creation of virtual environments for robot development using Isaac Sim, an advanced virtual environment known for realistic physics simulation, rich sensor simulation, and graphical rendering. 
The environment model, derived from terrain and trees of hierarchy maps, is seamlessly integrated (see Fig. \ref{fig: isaac_1}).

The first scenario simulates a car-like robot equipped with LiDAR operating in orchards (Fig. \ref{fig: isaac_1} (a)). The real-world terrain is converted into 2D images and meshes are constructed based on these images. Tree locations are imported directly from EPM predictions and tree models are generated by extracting meshes from the renderings.
In the second scenario (Fig. \ref{fig: isaac_1} (b) and (c)), fruit models with fragile joints are added to the trees, creating a reinforcement learning environment for robotic harvesting tasks. The harvesting robot, equipped with a UR-5 manipulator, a soft gripper and a camera on the remote robot base, is trained and evaluated in this virtual environment.

\section{Conclusion} \label{section: conclusion}
This study investigates the utilisation of robots in plant phenotyping to increase efficiency and reduce labour-intensive tasks. The proposed system consists of three key sub-systems that address environmental information processing, motion planning for data acquisition, and modelling using data collected by the robot.
Experimental results demonstrate the effectiveness of the system, particularly in outdoor environments with mild terrains, where the robot can collect high-quality samples leading to superior plant phenotypic models. 
For undulating terrains, typically challenging for plant phenotyping, our enhancements to the NeRF model also exhibit promise in generating quality plant phenotypic models.
Future research will focus on further refining phenotyping quality by exploiting the robotic system, incorporating the robotic arm, and making additional improvements to the NeRF model to minimise artefacts in challenging environments.

\bibliographystyle{IEEEtran}
\bibliography{root}
\end{document}